\documentclass[11pt,a4paper]{article}
\usepackage[hyperref]{acl2020}
\usepackage{times}
\usepackage{latexsym}

\usepackage{microtype}
\usepackage{amsmath}
\usepackage{graphicx}
\usepackage{subfigure}
\usepackage{ dsfont }
\usepackage{bbm}
\usepackage{hyperref}

\aclfinalcopy 
\def\aclpaperid{100} 

\title{What's The Latest? A Question-driven News Chatbot}

\author{Philippe Laban \\
  UC Berkeley\\
  \texttt{phillab@berkeley.edu} \\\And
  John Canny \\
  UC Berkeley\\
  \texttt{canny@berkeley.edu} \\\And
  Marti A. Hearst\\
  UC Berkeley\\
  \texttt{hearst@berkeley.edu}
  }

\date{}

\begin{document}
\maketitle
\begin{abstract}
    This work describes an automatic news chatbot that draws content from a diverse set of news articles and creates conversations with a user about the news. Key components of the system include the automatic organization of news articles into topical chatrooms, integration of automatically generated questions into the conversation, and a novel method for choosing which questions to present which avoids repetitive suggestions.
    We describe the algorithmic framework and present the results of a usability study that shows that news readers using the system successfully engage in multi-turn conversations about specific news stories.
\end{abstract}

\section{Introduction}

Chatbots offer the ability for interactive information access, which could be of great value in the news domain. As a user reads through news content, interaction could enable them to ask clarifying questions and go in depth on selected subjects. Current news chatbots have minimal capabilities, with content hand-crafted by members of news organizations, and cannot accept free-form questions.

To address this need, we design a new approach to interacting with large news collections. We designed, built, and evaluated a fully automated news chatbot that bases its content on a stream of news articles from a diverse set of English news sources. This in itself is a novel contribution. 

Our second contribution is with respect to the scoping of the chatbot conversation. 
The system organizes the news articles into chatrooms, each revolving around a \textit{story}, which is a set of automatically grouped news articles about a topic (e.g., articles related to Brexit).

The third contribution is a method to keep track of the state of the conversation to avoid repetition of information. For each news story, we first generate a set of essential questions and link each question with content that answers it. The motivating idea is: \textit{two pieces of content are redundant if they answer the same questions.} As the user reads content, the system tracks which questions are answered (directly or indirectly) with the content read so far, and which remain unanswered.
We evaluate the system through a usability study. 

The remainder of this paper is structured as follows. Section~\ref{section:system_description} describes the system and the content sources, Section~\ref{section:conversation_state} describes the algorithm for keeping track of the conversation state, Section~\ref{section:study_results} provides the results of a usability study evaluation and Section~\ref{section:related_work} presents relevant prior work.

The system is publicly available at \url{https://newslens.berkeley.edu/} and a demonstration video is available at this link: \url{https://www.youtube.com/watch?v=eze9hpEPUgo}.

\begin{figure*}[t]
    \centering
    \subfigure[Homepage]{%
        \label{fig:mobile_homepage}%
        \includegraphics[width=0.3\textwidth]{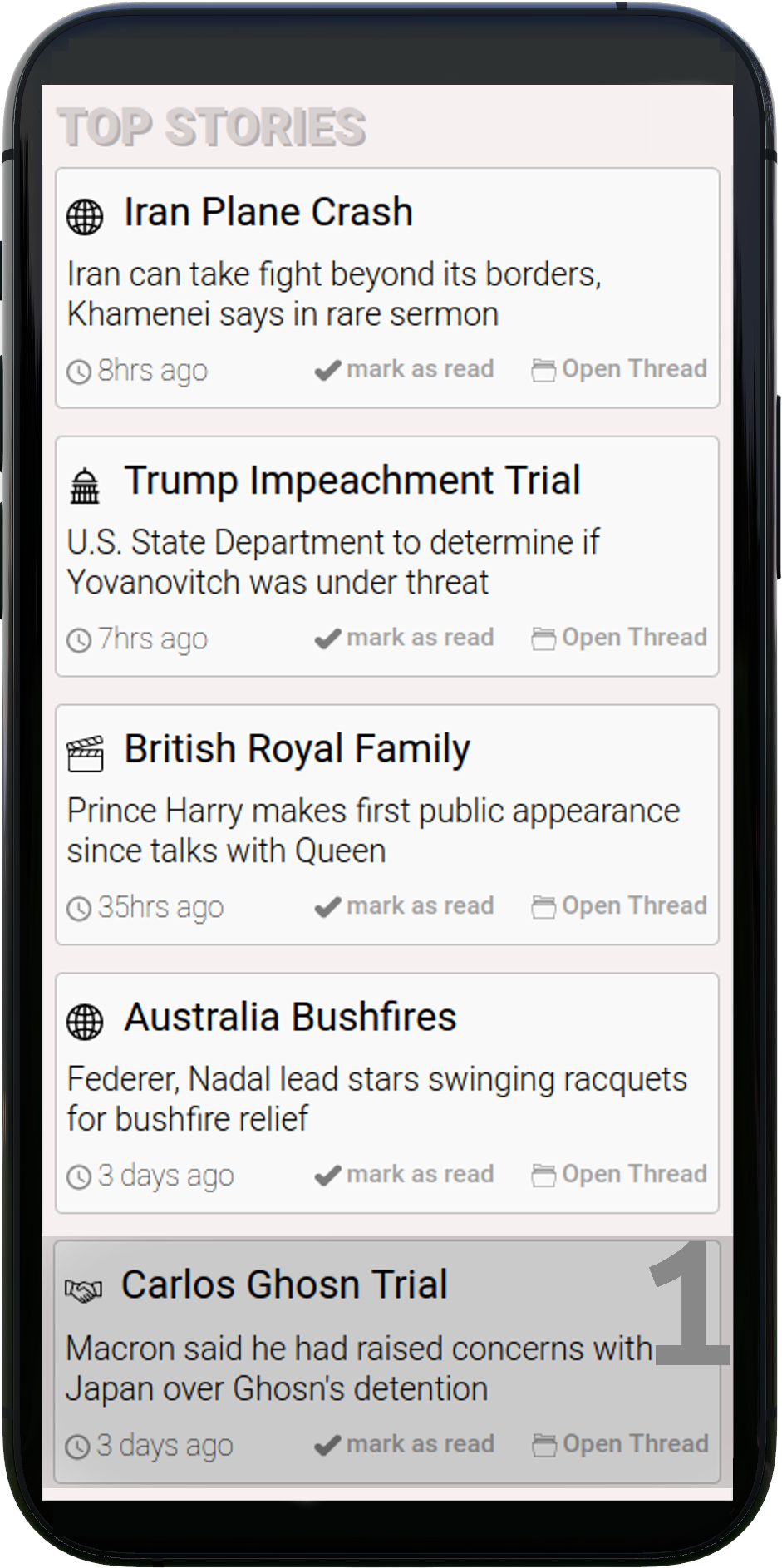}}%
    \qquad
    \subfigure[Initiating a Chatroom]{%
        \label{fig:mobile_conv_start}%
        \includegraphics[width=0.3\textwidth]{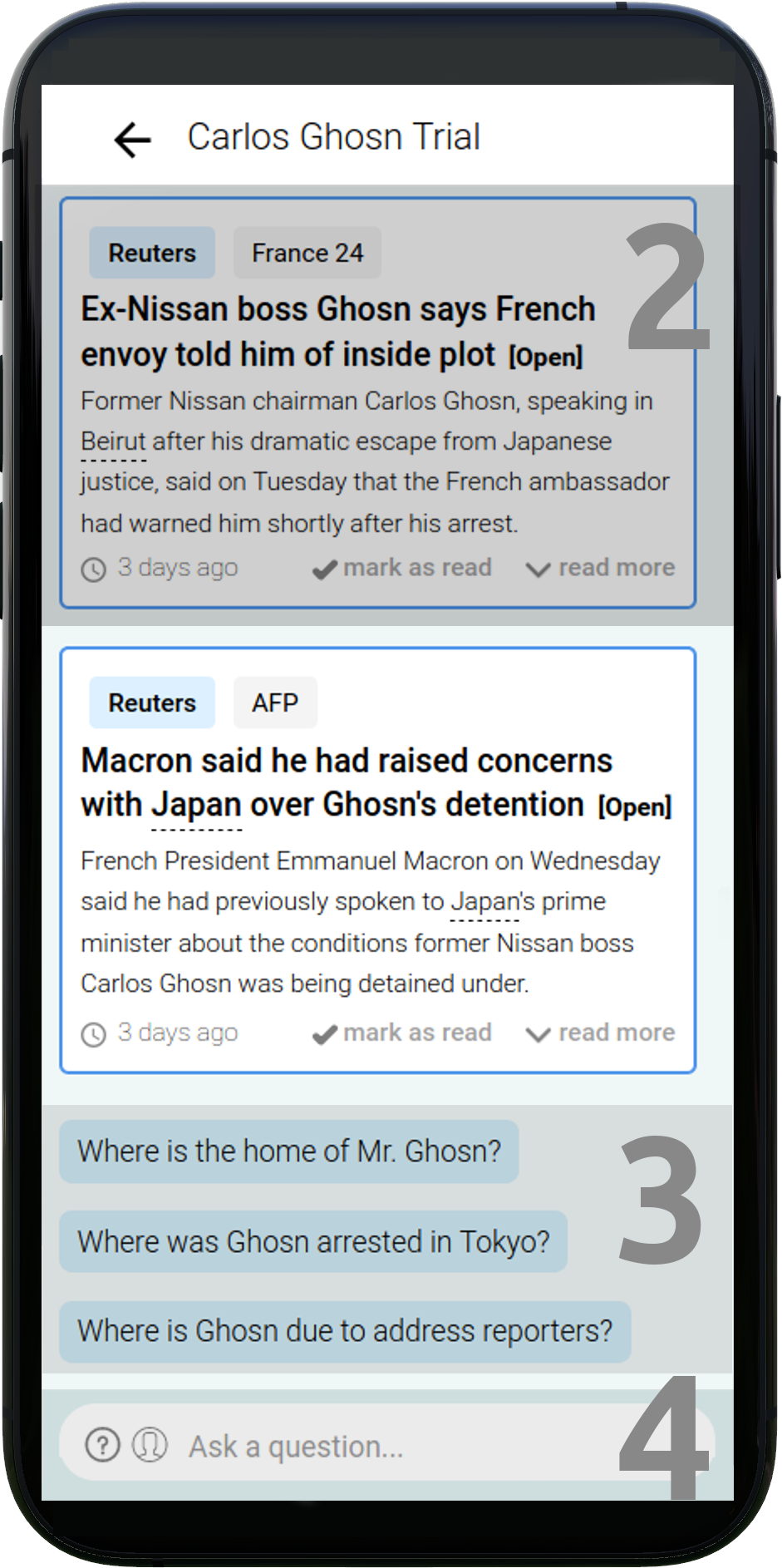}}%
    \qquad
    \subfigure[Chatroom Q\&A]{%
        \label{fig:mobile_qa}%
        \includegraphics[width=0.3\textwidth]{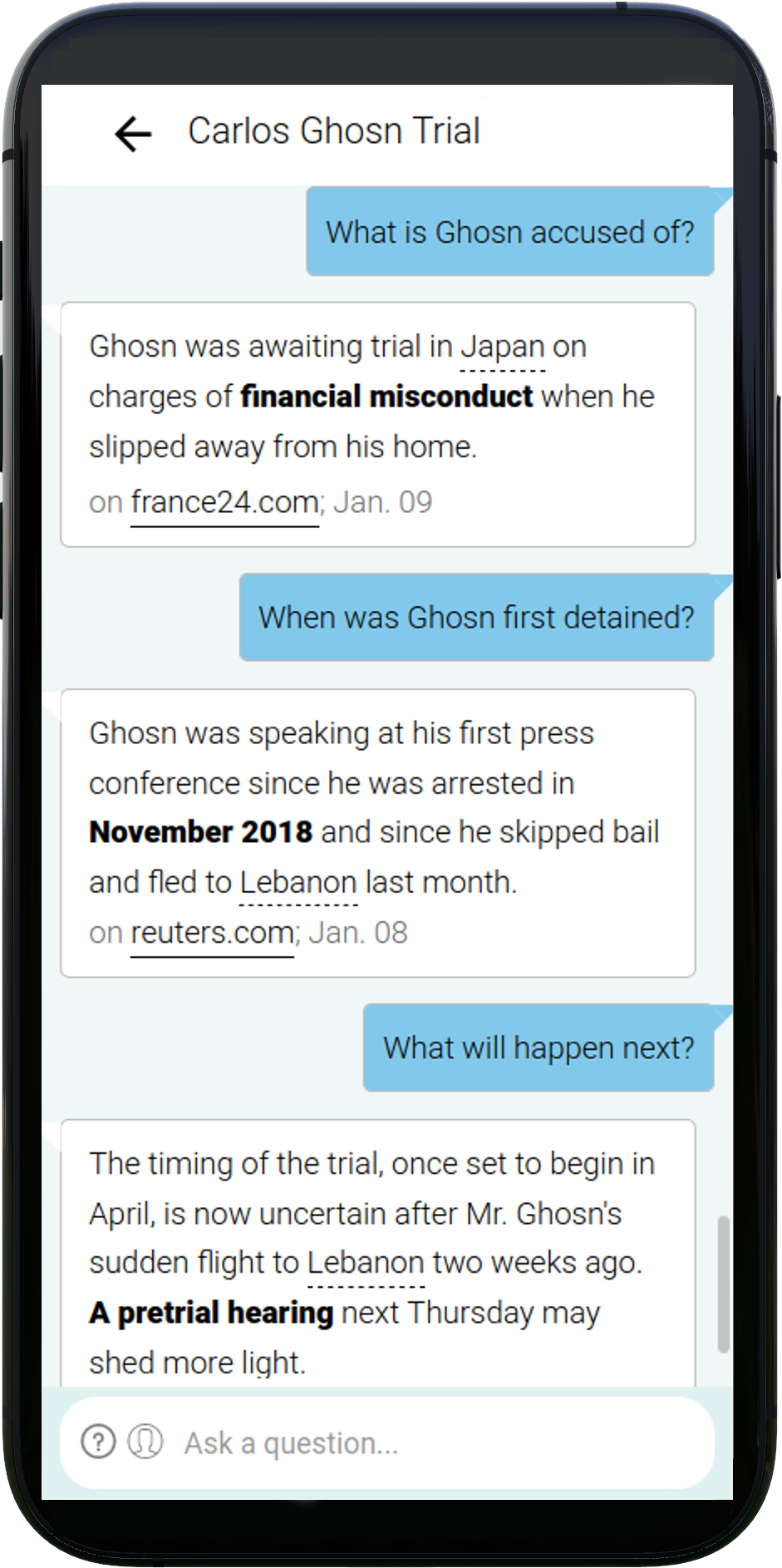}}%

    \caption{\textbf{Screenshots of the news chatbot} \subref{fig:mobile_homepage} Homepage lists most recently active chatrooms (Zone 1 is an example chatroom) \subref{fig:mobile_conv_start} Newly opened chatroom: Zone 2 is an event message, Zone 3 the Question Recommendation module, and Zone 4 a text input for user-initiated questions. Event messages are created via abstractive summarization. \subref{fig:mobile_qa} Conversation continuation with Q\&A examples. Sentences shown are extracted from original articles, whose sources are shown. Answers to questions are bolded.}
    \label{figure:mobile_interface}
\end{figure*}

\section{System Description}
\label{section:system_description}

This section describes the components of the chatbot: the content source,  the user interface, the supported user actions and the computed system answers. Appendix~\ref{appendix:resources} lists library and data resources used in the system.

\subsection{Content Sources}

We form the content for the chatbot from a set of news sources. We have collected an average of 2,000 news articles per day from 20 international news sources starting in 2010. The news articles are clustered into \textit{stories}: groups of news articles about a similar evolving topic, and each story is automatically named \cite{laban2017newslens}. Some of the top stories at the time of writing are shown in Figure~\ref{fig:mobile_homepage}.

\subsection{User Interface}
The chatbot supports information-seeking: the user is seeking information and the system delivers information in the form of news content. 

The homepage (Figure~\ref{fig:mobile_homepage}) lists the most active stories, and a user can select a story to enter its respective chatroom (Figure~\ref{fig:mobile_conv_start}). The separation into story-specific rooms achieves two objectives: (1) clarity to the user, 
as the chatrooms allow the user to exit and enter chatrooms to come back to conversations, and (2) limiting the scope of each dialogue is helpful from both a usability and a technical standpoint, as it helps reduce ambiguity and search scope. For example, answering a question like: ``What is the total cost to insurers so far?'' is easier when knowing the scope is the Australia Fires, compared to all of news.

Articles in a story are grouped into events, corresponding to an action that occurred in a particular time and place. For each event, the system forms an \textit{event message} by combining the event's news article headlines generated by an abstractive summarizer model \cite{laban2020summary_loop}.

Zone 2 in Figure~\ref{fig:mobile_conv_start} gives an example of an event message. The event messages form a chronological timeline in the story.

Because of the difference in respective roles, we expect user messages to be shorter than system responses, which we aim to be around 30 words.

\subsection{User Actions}

During the conversation, the user can choose among different kinds of actions.

\textbf{Explore the event timeline.} A chatroom conversation starts with the system showing the two most recent event messages of the story (Figure~\ref{fig:mobile_conv_start}). These messages give minimal context to the user necessary to start a conversation. When the event timeline holds more than two events, a ``See previous events'' button is added at the top of the conversation, allowing the user to go further back in the event timeline of the story.

\textbf{Clarify a concept.} The user can ask a clarification question regarding a person or organization (e.g., Who is Dennis Muilenburg?), a place (e.g., Where is Lebanon?) or an acronym (e.g., What does NATO stand for?). For a predetermined list of questions, the system will see if an appropriate Wikipedia entry exists, and will respond with the first two paragraphs of the Wikipedia page. For geographical entities, the system additionally responds with a geographic map when possible.

\textbf{Ask an open-ended question.} A text box (Zone 4 in Figure~\ref{fig:mobile_conv_start}) can be used to ask any free-form question about the story. A Q\&A system described in Section~\ref{section:conversation_state} attempts to find the answer in any paragraph of any news article of the story. If the Q\&A system reaches a confidence level about at least one paragraph containing an answer to the question, the chatbot system answers the question using one of the paragraphs. In the system reply the Q\&A selected answer is bolded. Figure~\ref{fig:mobile_qa} shows several Q\&A exchanges.

\textbf{Select a recommended question.} A list of three questions generated by the algorithm described in Section~\ref{section:conversation_state} is suggested to the user at the bottom of the conversation (Zone 3 in Figure~\ref{fig:mobile_conv_start}). Clicking on a recommended questions corresponds to asking the question in free-form. However, because recommended questions are known in advance, we pre-compute their answers to minimize user waiting times.

\section{Conversation State}
\label{section:conversation_state}
\begin{figure}
    \centering
    \includegraphics[width=0.47\textwidth]{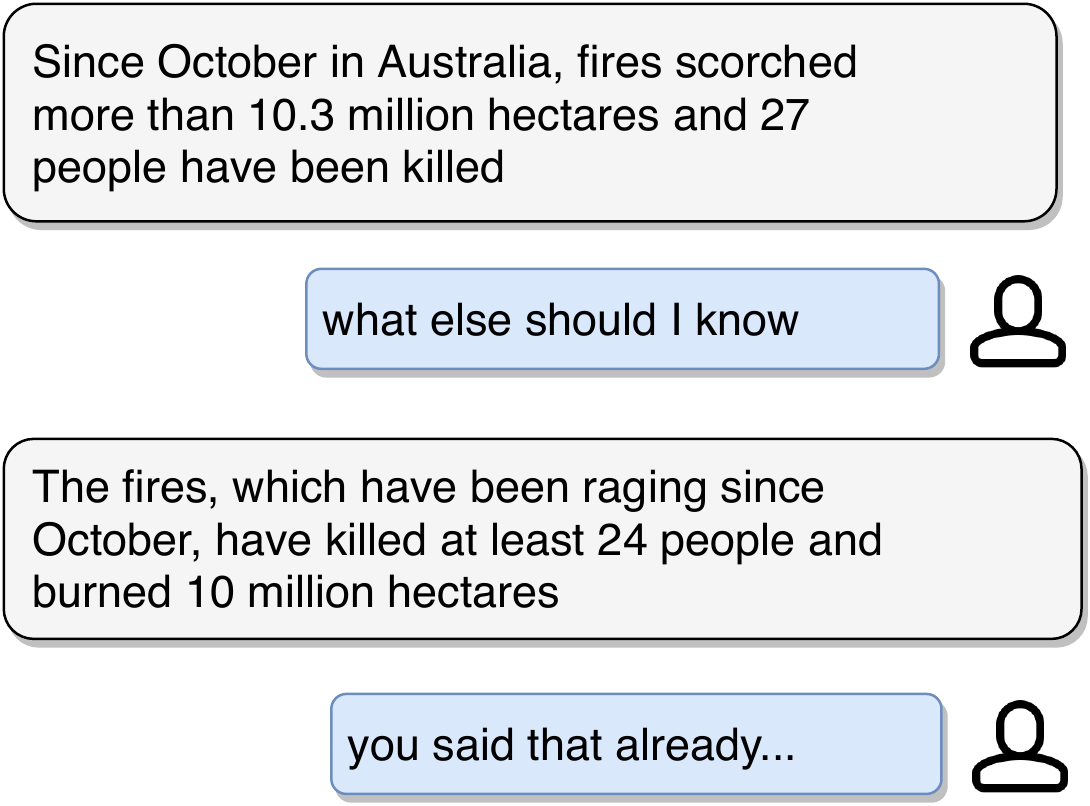}
    \caption{\textbf{Example of repetition from the system.} Repeating facts with different language is undesirable in a news chatbot. We introduce a novel question tracking method that attempts to minimize repetition.}
    \label{fig:repetition_problem}
\end{figure}

\begin{figure}
    \centering
    \includegraphics[width=0.47\textwidth]{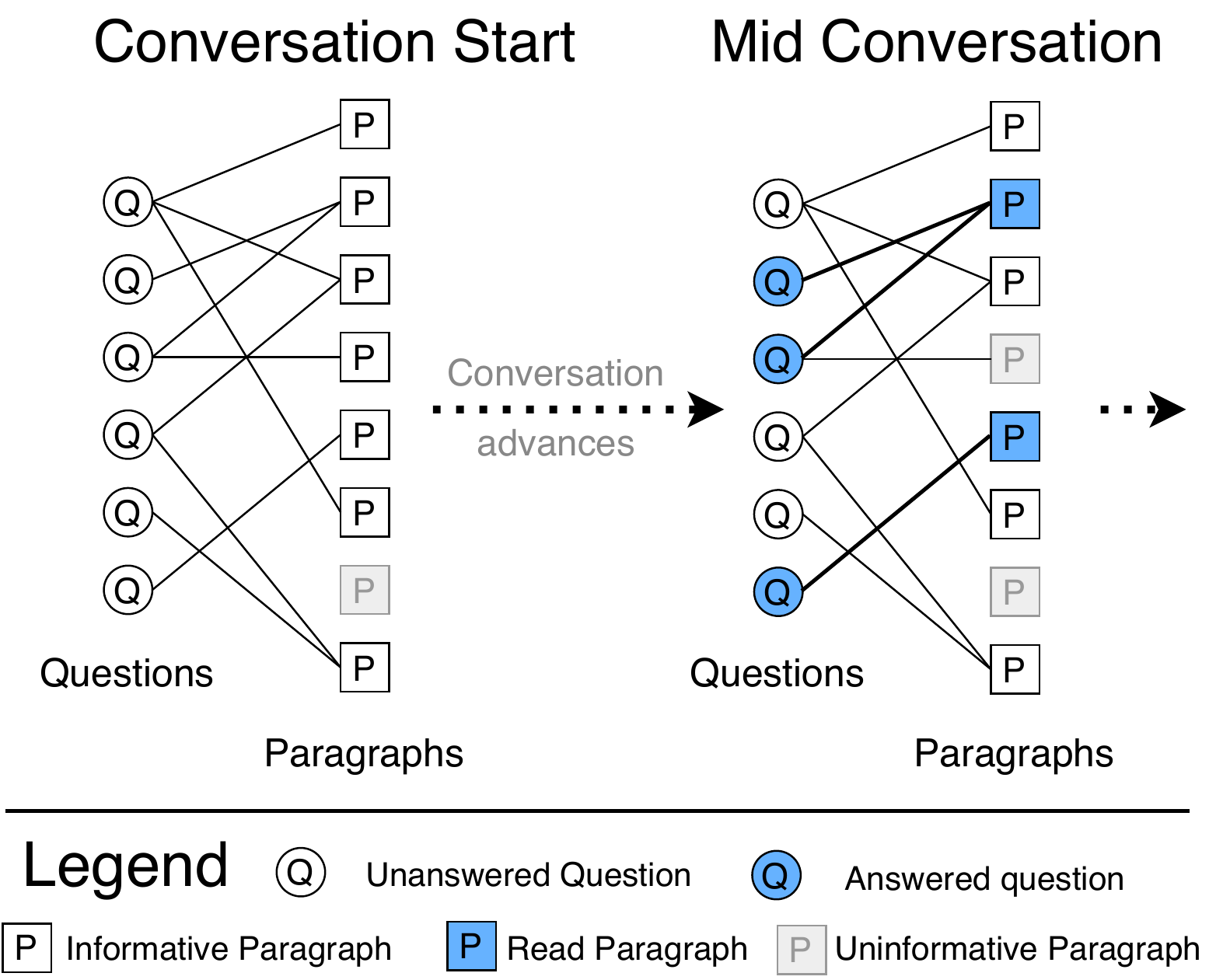}
    \caption{\textbf{Conversation state is tracked with the P/Q graph.} As the conversation advances, the system keeps track of answered questions. Any paragraph that does not answer a new question is discarded. Questions that are not answered yet are recommended.}
    \label{fig:conversation_state}
\end{figure}

One key problem in dialogue systems is that of keeping track of conveyed information, and avoiding repetition in system replies (see example in Figure~\ref{fig:repetition_problem}). This problem is amplified in the news setting, where different news organizations cover content redundantly.

We propose a solution that takes advantage of a Question and Answer (Q\&A) system. As noted above, the motivating idea is that two pieces of content are redundant if they answer the same questions. In the example of Figure~\ref{fig:repetition_problem}, both system messages answer the same set of questions, namely: ``When did the fires start?'', ``How many people have died?'' and ``How many hectares have burned?'', and can therefore be considered redundant.

Our procedure to track the knowledge state of a news conversation consists of the following steps: (1) generate candidate questions spanning the knowledge in the story, (2) build a graph connecting paragraphs with questions they answer, (3) during a conversation, use the graph to track what questions have been answered already, and avoid using paragraphs that do not answer new questions. 

\textbf{Question Candidate Generation.} We fine-tune a GPT2 language model \cite{radford2019language} on the task of question generation using the SQuAD 2.0 dataset \cite{rajpurkar2018know}. At training, the model reads a paragraph from the training set, and learns to generate a question associated with the paragraph. For each paragraph in each article of the story (the paragraph set), we use beam search to generate $K$ candidate questions. In our experience, using a large beam size ($K$=20) is important, as one paragraph can yield several valid questions.  Beam search enforces exploration, with the first step of beam search often containing several interrogative words (what, where...). 

For a given paragraph, we reduce the set of questions by deduplicating questions that are lexically close (differ by at most 2 words), and removing questions that are too long ($>$12 words) or too short ($<$5 words).

\textbf{Building the P/Q graph.} We train a standard Q\&A model, a Roberta model \cite{liu2019roberta} finetuned on SQuAD 2.0 \cite{rajpurkar2018know}, and use this model to build a paragraph / question bipartite graph (P/Q graph). In the P/Q graph, we connect any paragraph (P node), with a question (Q node), if the Q\&A model is confident that paragraph P answers question Q. An example bipartite graph obtained is illustrated in Figure~\ref{fig:conversation_state}, with the question set on the left, the paragraph set on the right, and edges between them representing model confidence about the answer.

Because we used a large beam-size when generating the questions, we perform a pruning step on the questions set. Our pruning procedure is based on the realization that two questions are redundant if they connect to the same subset of paragraphs (they cover the same content). Our objective is to find the smallest set of questions that cover all paragraphs. This problem can be formulated as a standard graph theory problem known as the set cover problem, and we use a standard heuristic algorithm \cite{caprara1999heuristic}.
After pruning, we obtain a final P/Q graph, a subgraph of the original consisting only of the covering set questions.

The P/Q graph embodies interesting properties. First, the degree of a question node measures how often a question is answered by distinct paragraphs, providing a measure of the question's importance to the story. The degree of a paragraph node indicates how many distinct questions it answers, an estimate of its relevance to a potential reader. Finally, the graph can be used to measure question relatedness: if two questions have non-empty neighboring sets (i.e., some paragraphs answer both questions), they are likely to be related questions, which can be used as a way to suggest follow-up questions. 

\textbf{Using the P/Q graph.}
At the start of a  conversation, no question is answered, since no paragraph has been shown to the user. Therefore, the system initializes a blank P/Q graph (left graph in Figure~\ref{fig:conversation_state}). As the system reveals paragraphs in the conversation, they are marked as \textit{read} in the P/Q graph (shaded blue paragraphs in the right graph of Figure~\ref{fig:conversation_state}). According to our Q\&A model, any question connected to a read paragraph is \textit{answered}, so we mark all neighbors of read paragraphs as answered questions (shaded blue questions on the right graph of Figure~\ref{fig:conversation_state}). At any stage in the conversation, if a paragraph is connected to only answered questions, it is deemed \textit{uninformative}, as it will not reveal the answer to a new question.

As the conversation moves along, more paragraphs are read, increasing the number of answered questions, which in turn, increases the number of uninformative paragraphs. We program the system to prioritize paragraphs that answer the most unanswered questions, and disregard uninformative paragraphs. We further use the P/Q graph to recommend questions to the user. We select unanswered questions and prioritize questions connected to more unread paragraphs, recommending questions three at a time.

\section{Study Results}
\label{section:study_results}
We conducted a usability study in which participants were assigned randomly to one of three configurations:
\begin{itemize}
\item TOPQR: the recommended questions are the most informative according to the algorithm in Section~\ref{section:conversation_state} (N=18), 

\item RANDQR: the recommended questions are randomly sampled from the questions TOPQR would not select (however, near duplicates will appear in this set) (N=16), 

\item NOQR: No questions are recommended, and the Question Recommendation module (Zone 3 in Figure~\ref{fig:mobile_conv_start}) is hidden (N=22).
\end{itemize}

These are contrasted in order to test (a) if showing automatically generated questions is beneficial to news readers, and (b) to assess the question tracking algorithm against a similar question recommendation method with no conversation state.

\subsection{Study Setup}

We used Amazon Mechanical Turk to recruit participants, restricting the task to workers in English-speaking countries having previous completed 1500 tasks (HITs) and an acceptance rate of at least 97\%. Each participant was paid a flat rate of \$2.50 with the study lasting a total of 15 minutes.
During the study, the participants first walked through an introduction to the system, then read the news for 8 minutes, and finally completed a short survey.

During the eight minutes of news reading, participants were requested to select at least 2 stories to read from a list of the 20 most recently active news stories.\footnote{We manually removed news stories that were predominantly about politics, to avoid heated political questions, which were not under study here.} The participants were prompted to choose stories they were interested in.

The survey consisted of two sections: a satisfaction section, and a section for general free-form feedback.
The satisfaction of the participants was surveyed using the standard Questionnaire for User Interaction Satisfaction (QUIS) \cite{norman1998questionnaire}. QUIS is a series of questions about the usability of the system (ease of use, learning curve, error messages clearness, etc.) answered on a 7-point Likert scale. We modify QUIS by adding two questions regarding questions and answers: ``Are suggested questions clear?'' and ``Are answers to questions informative?''
A total of fifty-six participants completed the study. We report on the usage of the system, the QUIS Satisfaction results and  textual comments  from the participants.

\subsection{Usage statistics}

\begin{table}[]
    \resizebox{0.5\textwidth}{!}{%
    \begin{tabular}{llll}
     \hline
     \textbf{Measured Value} & \textbf{TOPQR} & \textbf{RANDQR} & \textbf{NOQR} \\ \hline
     \# participants & 18 & 16 & 22 \\
     \# chatrooms opened & 3.2 & 2.9 & 3.1 \\
     \# msgs. / chatroom & 24.9 $^*$ & 15.3  $^*$ & 8.1 \\\hline
     \# rec. questions asked & 11.9 $^*$ & 8.2  $^*$ & - \\
     \# own questions asked & 1.5 & 1.1 & 2.2 \\
     \# total questions asked & 13.4 $^*$ & 9.3  $^*$ & 2.2 \\ \hline
      latency (seconds) & 1.84 $^*$ & 1.88  $^*$ & 4.51 \\ \hline
    \end{tabular}
    }
    \caption{\textbf{Usage statistics of the news chatbot during the usability study.} Participants either saw most informative recommended questions (TOPQR), randomly selected recommended questions (RANDQR) or no recommended questions (NOQR). $*$ signifies statistical difference with NOQR (p $<$ 0.05).}
    \label{table:usage}
\end{table}

We observed that participants in the QR-enabled interfaces (TOPQR and RANDQR) had longer conversations than the NOQR setting, with an average chatroom conversation length of 24.9 messages in the TOPQR setting. Even though the TOPQR setting had average conversation length longer than RANDQR, this was not statistically significant.

This increase in conversation length is mostly due to the use of recommended questions, which are convenient to click on. Indeed, users clicked on 8.2 questions on average in RANDQR and 11.9 in TOPQR. NOQR participants wrote on average 2.2 of their own questions, which was not statistically higher than TOPQR (1.5) and RANDQR (1.1), showing that seeing recommended questions did not prevent participants from asking their own questions.

When measuring the latency of system answers to participant questions, we observe that the average wait time in TOPQR (1.84 seconds) and RANDQR (1.88 seconds) settings is significantly lower than NOQR (4.51 seconds). This speedup is due to our ability to pre-compute answers to recommended questions, an additional benefit of the QR graph pre-computation.

\subsection{QUIS Satisfaction Scores}

Overall, the systems with question recommendation enabled (TOPQR and RANDQR) obtained higher average satisfaction on most measures than the NOQR setting. That said, statistical significance was only observed in 4 cases between  TOPQR and NOQR, with participants judging the TOPQR interface to be more stimulating and satisfying.

Although not statistically significant, participants rated the suggested questions for TOPQR almost 1 point higher than RANDQR, providing some evidence that incorporating past viewed information into question selection is beneficial.

Participants judged the answers to be more informative in the TOPQR setting. We interpret this as evidence that the QR module helps teach users what types of questions the system can answer, enabling them to get better answers. Several NOQR participants asked ``What can I ask?'' or equivalent.

\begin{table}[]
    \resizebox{0.48\textwidth}{!}{%
    \begin{tabular}{lccc}
    \hline
    \textbf{Measured Value} & \textbf{TOPQR} & \textbf{RANDQR} & \textbf{NOQR} \\ \hline
    (1) dull ... stimulating (7) & 5.28 $^*$ & 5.06 & 4.20 \\
    (1) frustrating ... satisfying (7) & 5.00 $^*$ & 4.43 & 4.00 \\
    (1) rigid ... flexible (7) & 4.71 & 4.66 & 4.14 \\
    (1) terrible ... wonderful (7) & 4.79 & 4.69 & 4.20 \\ \hline
    exploring new features & 5.80 & 5.50 & 5.14 \\
    learning to operate & 5.40 & 5.25 & 5.06 \\
    \begin{tabular}[c]{@{}l@{}}performing task is\\ straightforward\end{tabular} & 5.40 & 5.56 & 5.20 \\ \hline
    system reliability & 5.80 & 5.19 & 5.67 \\
    system speed & 6.20 & 5.87 & 5.44 \\ \hline
    rec. questions are clear & 5.78 $^*$ & 4.87 & 4.28 \\
    answers are informative & 5.07 $^*$ & 4.44 & 3.64 \\ \hline
    \end{tabular}
    }
    \caption{\textbf{QUIS satisfaction results.}  Likert values on a scale from 1 to 7, higher is better unless stated otherwise.
    $*$ signifies statistical difference with NOQR (p $<$ 0.05).}
    \label{table:satisfaction}
\end{table}

\subsection{Qualitative Feedback}

Thirty-four of the fifty-six participants opted to give general feedback via an open ended text box. We tagged the responses into major themes:
\begin{enumerate}
	\item 19 participants (7 TOPQR, 7 RANDQR, 5 NOQR) expressed interest in the system (e.g., \textit{I enjoyed trying this system out. I particularly liked that stories are drawn from various sources.})
	\item 11 participants (4, 3, 4) mentioned the system did not correctly reply to questions asked (e.g., \textit{Some of the questions kind of weren't answered exactly, especially in the libya article}), 
	\item 10 participants (2, 3, 5) found an aspect of the interface confusing (e.g., \textit{This system has potential, but as of right now it seems too overloaded and hard to sort through.})
	\item 6 participants (4, 2, 0) thought the questions were useful (e.g., \textit{I especially like the questions at the bottom. Sometimes it helps to remember some basic facts or deepen your understanding})
\end{enumerate}

The most commonly mentioned limitation was Q\&A related errors, a limitation we hope to mitigate as automated Q\&A continues progressing.

\section{Related Work}
\label{section:related_work}
\textbf{News Chatbots.} Several news agencies have ventured in the space of dialogue interfaces as a way to attract new audiences. The chatbots are often manually curated for the dialogue medium and advanced NLP machinery such as a Q\&A systems are not incorporated into the chatbot.

On BBC's Messenger chatbot\footnote{https://www.messenger.com/t/BBCPolitics}, a user can enter search queries, such as ``latest news'' or ``Brexit news'' and obtain a list of latest BBC articles matching the search criteria. In the chatbot produced by Quartz\footnote{https://www.messenger.com/t/quartznews}, journalists hand-craft news stories in the form of pre-written dialogues (aka choose-your-own adventure). At each turn, the user can choose from a list of replies, deciding which track of the dialogue-article is followed. CNN\footnote{https://www.messenger.com/t/cnn} has also experimented with choose-your-own adventure articles, with the added ability for small talk.

\textbf{Relevant Q\&A datasets.} NewsQA \cite{trischler2017newsqa} collected a dataset by having a crowd-worker read the summary of a news article and ask a follow-up question. Subsequent crowd-workers answered the question or marked it as not-answerable. NewsQA's objective was to collect a dataset, and we focus on building a usable dialogue interface for the news with a Q\&A component.

CoQA \cite{reddy2019coqa} and Quac \cite{choi2018quac} are two datasets collected for questions answering in the context of a dialogue. For both datasets, two crowd-workers (a student and a teacher) have a conversation about a piece of text (hidden to the student in Quac). The student \textit{must ask} questions of the teacher, and the teacher answers using extracts of the document. In our system, the questions asked by the user are answered automatically, introducing potential errors, and the user can choose to ask questions or not.

In this work, the focus is not on the collection of naturally occurring questions, but in putting a Q\&A system in use in a news dialogue system, and observing the extent of its use.

\textbf{Question Generation} (QG) has become an active area for text generation. A common approach is to use a sequence to sequence model \cite{du2017learning}, encoding the paragraph (or context), an optional target answer (answer-aware \cite{sun2018answer}), and decoding a paired question. This common approach focuses on the generation of a single article, from a single piece of context, often a paragraph. We argue that our framing of the QG problem as the generation of a series of questions spanning several (possibly redundant) documents is a novel task.

\citet{Krishna2019GeneratingQH} build a hierarchy of questions generated for a single document; the document is then reorganized into a ``Squashed'' document, where paragraphs and questions are interleaved. Because our approach is based on using multiple documents as the source, compiling all questions into a single document would be long to read, so we opt for a chatbot.

\section{Discussion}

During the usability study, we obtained direct and indirect feedback from our users, and we summarize limitations that could be addressed in the system.

\textbf{Inability to Handle Small Talk}. 4 participants attempted to have small talk with the chatbot (e.g. asking ``how are you''). The system most often responded inadequately, saying it did not understand the request.
Future work may include gently directing users who engage in small talk to a chit-chat-style interface.

\textbf{Inaccurate Q\&A system}. 32\% of the participants mentioned that answers are often off-track or irrelevant. This suggests that further improvements in Q\&A systems are needed.

\textbf{Dealing with errors}. Within the current framework, errors are bound to happen, and easing the user's path to recovery could improve the user experience. 
\section{Conclusion}

We presented a fully automated news chatbot system, which leverages an average of 2,000 news articles a day from a diverse set of sources to build chatrooms for important news stories. In each room, the system takes note of generated questions that have already been answered, to minimize repetition of information to the news reader.

A usability study reveals that when the chatbot recommends questions, news readers tend to have longer conversations, with an average of 24 messages exchanged. These conversation consist of combination of recommended and user-created questions.

\section*{Acknowledgments}

We would like to thank Ruchir Baronia for early prototyping and the ACL reviewers for their helpful comments. This work was supported by a Bloomberg Data Science grant. We also gratefully acknowledge support received from an Amazon Web Services Machine Learning Research Award and an NVIDIA Corporation GPU grant.

\bibliography{anthology,acl2020}
\bibliographystyle{acl_natbib}

\clearpage
\appendix
\renewcommand{\thetable}{A\arabic{table}}
\setcounter{table}{0}
\renewcommand{\thefigure}{A\arabic{figure}}
\setcounter{figure}{0}

\section{Resources Used}
\label{appendix:resources}
 The libraries and data sources used in the described system are as follows:

\textbf{Transformers library}\footnote{https://github.com/huggingface/transformers} used to train the GPT2-based Question Generation model and the Roberta-based Q\&A model.

\textbf{spaCy library}\footnote{https://github.com/explosion/spaCy} used to do named-entity extraction, phrase and keyword extraction. 

\textbf{Wikidata}\footnote{https://www.wikidata.org/} for entity linking and collection of textual content of relevant Wikipedia pages used in special case questions.

\textbf{MongoDB}\footnote{https://www.mongodb.com/} and \textbf{Flask}\footnote{https://flask.palletsprojects.com/en/1.1.x/} for storing and serving the content to the user.

\textbf{SetCoverPy}\footnote{https://github.com/guangtunbenzhu/SetCoverPy} for its implementation of standard set cover algorithms in Python.

\textbf{List of news sources} present in the dataset used by the system, in alphabetical order: Aa.com.tr, Afp.com, Aljazeera.com, Allafrica.com, Apnews.com, Bbc.co.uk, Bloomberg.com, Chicagotribune.com, Chinadaily.com.cn, Cnet.com, Cnn.com, Foxnews.com, France24.com, Independent.co.uk, Indiatimes.com, Latimes.com, Mercopress.com, Middleeasteye.net, Nytimes.com, Reuters.com, Rt.com, Techcrunch.com, Telegraph.co.uk, Theguardian.com, Washingtonpost.com

\end{document}